\documentclass{article} % For LaTeX2e
\usepackage{iclr2016_conference,times}
\usepackage{multirow}
\usepackage{courier}
\usepackage{array}
\usepackage{url}
\usepackage{color}
\usepackage{mathrsfs}
\usepackage{todonotes}
\usepackage[TABBOTCAP]{subfigure}
\usepackage{capt-of}
\usepackage{graphicx, caption}
\usepackage{booktabs}
\usepackage{floatrow}
\usepackage{changepage}
\usepackage[bottom]{footmisc}
\usepackage{algorithm,algorithmic}
\usepackage{listings}
\usepackage[title]{appendix}
\newcommand\modelname{Magnet Loss}
\newcommand\modelnameshort{Magnet}

\usepackage{macros}

\usepackage[colorlinks,citecolor=blue,urlcolor=blue]{hyperref}

\iclrfinalcopy

\makeatletter
\newcommand\footnoteref[1]{\protected@xdef\@thefnmark{\ref{#1}}\@footnotemark}
\makeatother

\title{Metric Learning with\\ Adaptive Density Discrimination}

\author{Oren Rippel\\
MIT, Facebook AI Research\\
\texttt{rippel@math.mit.edu} \\
\And
\hspace{-2.3in}Manohar Paluri \\
\hspace{-2.3in}Facebook AI Research\\
\hspace{-2.3in}\texttt{mano@fb.com} \\
\AND
Piotr Dollar\\
Facebook AI Research\\
\texttt{pdollar@fb.com} \\
\And
\hspace{1.68in} Lubomir Bourdev\\
\hspace{1.68in} UC Berkeley\\
\hspace{1.68in} \texttt{lubomir.bourdev@gmail.com} \\
}

\newlength{\offsetleft}
\newlength{\offsetright}
\newenvironment{widepage}{\begin{adjustwidth}{-\offsetleft}{-\offsetright}%
    \addtolength{\textwidth}{\offsetleft+\offsetright}}%
{\end{adjustwidth}}

\begin{document}

\maketitle
%%%%%%%%%%%%%%%%%%%%%%%%%%%%%%%%%%%%%%%%%%%
\begin{abstract}
Distance metric learning (DML) approaches learn a transformation to a representation space where distance is in correspondence with a predefined notion of similarity. While such models offer a number of compelling benefits, it has been difficult for these to compete with modern classification algorithms in performance and even in feature extraction.

In this work, we propose a novel approach explicitly designed to address a number of subtle yet important issues which have stymied earlier DML algorithms. It maintains an explicit model of the distributions of the different classes in representation space. It then employs this knowledge to adaptively assess similarity, and achieve local discrimination by penalizing class distribution overlap.

We demonstrate the effectiveness of this idea on several tasks. Our approach achieves state-of-the-art classification results on a number of fine-grained visual recognition datasets, surpassing the standard softmax classifier and outperforming triplet loss by a relative margin of 30-40\%. In terms of computational performance, it alleviates training inefficiencies in the traditional triplet loss, reaching the same error in 5-30 times fewer iterations. Beyond classification, we further validate the saliency of the learnt representations via their attribute concentration and hierarchy recovery properties, achieving 10-25\% relative gains on the softmax classifier and 25-50\% on triplet loss in these tasks.
\end{abstract}

%%%%%%%%%%%%%%%%%%%%%%%%%%%%%%%%%%%%%%%%%%%
\section{Introduction}
The problem of classification is a mainstay task in machine learning, as it provides us with a coherent metric to gauge progress and juxtapose new ideas against existing approaches. To tackle various other tasks beyond categorization, we often require alternative representations of our inputs which provide succinct summaries of relevant characteristics. Here, classification algorithms often serve as convenient feature extractors: a very popular approach involves training a network for classification on a large dataset, and retaining the outputs of the last layer as inputs transferred to other tasks \citep{Donahue_decaf:a,Razavian_2014_CVPR_Workshops,Qian_2015_CVPR,snoek-etal-2015a}.

However, this paradigm exhibits an intrinsic discrepancy: we have no guarantee that our extracted features are suitable for any task but the particular classification problem from which they were derived. On the contrary: in our classification procedure, we propagate high-dimensional inputs through a complex pipeline, and map each to a single, scalar prediction. That is, we explicitly demand our algorithm to, ultimately, dispose of all information but class label. In the process, we destroy intra- and inter-class variation that would in fact be desirable to maintain in our features. 

In principle, we have no reason to compromise: we should be able to construct a representation which is amenable to classification, while still maintaining more fine-grained information. This philosophy motivates the class of distance metric learning (DML) approaches, which learn a transformation to a representation space where distance is in correspondence with a notion of similarity. Metric learning offers a number of benefits: for example, it enables zero-shot learning \citep{mensink13pami,Chopra:2005:LSM:1068507.1068961}, visualization of high-dimensional data \citep{maaten2008visualizing}, learning invariant maps \citep{Hadsell06dimensionalityreduction}, and graceful scaling to instances with millions of classes \citep{Schroff_2015_CVPR}. In spite of this, it has been difficult for DML-based approaches to compete with modern classification algorithms in performance and even in feature extraction.

Admittedly, however, these are two sides of the same coin: a more salient representation should, in theory, enable improved classification performance and features for task transfer. In this work, we strive to reconcile this gap. We introduce \emph{\modelname{}}, a novel approach explicitly designed to address subtle yet important issues which have hindered the quality of learnt representations and the training efficiency of a class of DML approaches. In essence, instead of penalizing individual examples or triplets, it maintains an explicit model of the distributions of the different classes in representation space. It then employs this knowledge to adaptively assess similarity, and achieve discrimination by reducing local distribution overlap. It utilizes clustering techniques to simultaneously tackle a number of components in model design, from capturing the distributions of the different classes to hard negative mining. For a particular set of assumptions in its configuration, it reduces to the familiar triplet loss \citep{Weinberger:2009:DML:1577069.1577078}.

We demonstrate the effectiveness of this idea on several tasks. Using a soft $k$-nearest-cluster metric for evaluation, this approach achieves state-of-the-art classification results on a number of fine-grained visual recognition datasets, surpassing the standard softmax classifier and outperforming triplet loss by a relative margin of 30-40\%. In terms of computational performance, it alleviates several training inefficiencies in traditional triplet-based approaches, reaching the same error in 5-30 times fewer iterations. Beyond classification, we further validate the saliency of the learnt representations via their attribute concentration and hierarchy recovery properties, achieving 10-25\% relative gains on the softmax classifier and 25-50\% on triplet loss in these tasks. 

\begin{figure*}[t]
\begin{widepage}
\centering%
\includegraphics[height=1.7in,trim=0cm 0cm 0cm 0cm,clip]{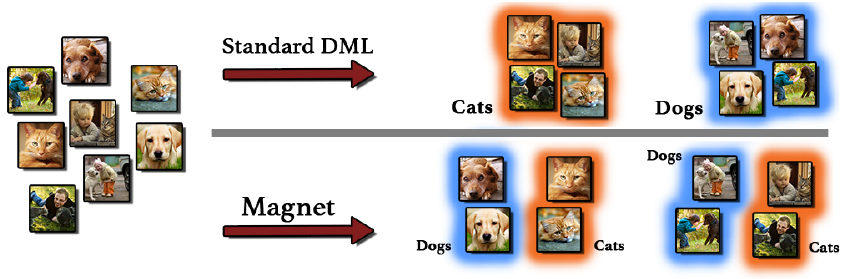}%
\caption{Distance metric learning approaches sculpt a representation space where distance is in correspondence with a notion of similarity. Traditionally, similarity is specified \emph{a-priori} and often strictly semantically. In contrast, \modelname{} adaptively sculpts its representation space by autonomously identifying and respecting intra-class variation and inter-class similarity.}
\label{fig:intro}
\end{widepage}
\end{figure*}

%%%%%%%%%%%%%%%%%%%%%%%%%%%%%%%%%%%%%%%%%%%%%%%%%%%%%%%%%%%%%%%%%%%%%%%%
\section{Motivation: Challenges in Distance Metric Learning}\label{challenges}
We start by providing an overview of challenges which we believe have been impeding the success of existing distance metric learning approaches. These will motivate our work to follow.

%%%%%%%%%%%%%%%%%%%%%%%%
\paragraph{Issue \#1: predefined target neighbourhood structure}\label{variation}
All metric learning approaches must define a relationship between similarity and distance, which prescribes neighbourhood structure. The corresponding training algorithm, then, learns a transformation to a representation space where this property is obeyed. In existing approaches, similarity has been canonically defined \emph{a-priori} by integrating available supervised knowledge. The most common is semantic, informed by class labels. Finer assignment of neighbourhood structure is enabled with access to additional prior information, such as similarity ranking \citep{wang2014learning} and hierarchical class taxonomy \citep{verma2012learning}. 

In practice, however, the only available supervision is often in the form of class labels. In this case, a ubiquitous solution is to enforce semantic similarity: examples of each class are demanded to be tightly clustered together, far from examples of other classes (for example, \citet{Schroff_2015_CVPR,NIPS2012_4808,NIPS2005_2947,Chopra:2005:LSM:1068507.1068961}). However, this collapses intra-class variation and does not embrace shared structure between different classes. Hence, this imposes too strong of a requirement, as each class is assumed to be captured by \emph{a single mode}.

This issue is well-known, and has motivated the notion of \emph{local} similarity: each example is designated only a small number of target neighbours of the same class \citep{Weinberger:2009:DML:1577069.1577078,Qian_2015_CVPR,Hadsell06dimensionalityreduction}. In existing work, these target neighbours are determined \emph{prior to training}: they are retrieved based on distances in the \emph{original input space}, and after which are never updated again. Ironically, this is in contradiction with our fundamental assumption which motivated us to pursue a DML approach in the first place. Namely, we want to learn a metric because we cannot trust distances in our original input space --- but on the other hand define target similarity using this exact metric that cannot be trusted! Thus, although this approach has the good intentions of encoding similarity into our representation, it harms intra-class variation and inter-class similarity by enforcing unreasonable proximity relationships. Apart from its information preservation ramifications, achieving predefined separation requires significant effort, which results in inefficiencies during training time.

\begin{figure*}[t]
\begin{widepage}
\centering
\begin{minipage}{0.9\textwidth}
\captionsetup{width=\textwidth}
\centering%
\includegraphics[width=\textwidth,trim=0cm 0cm 0cm 0cm,clip]{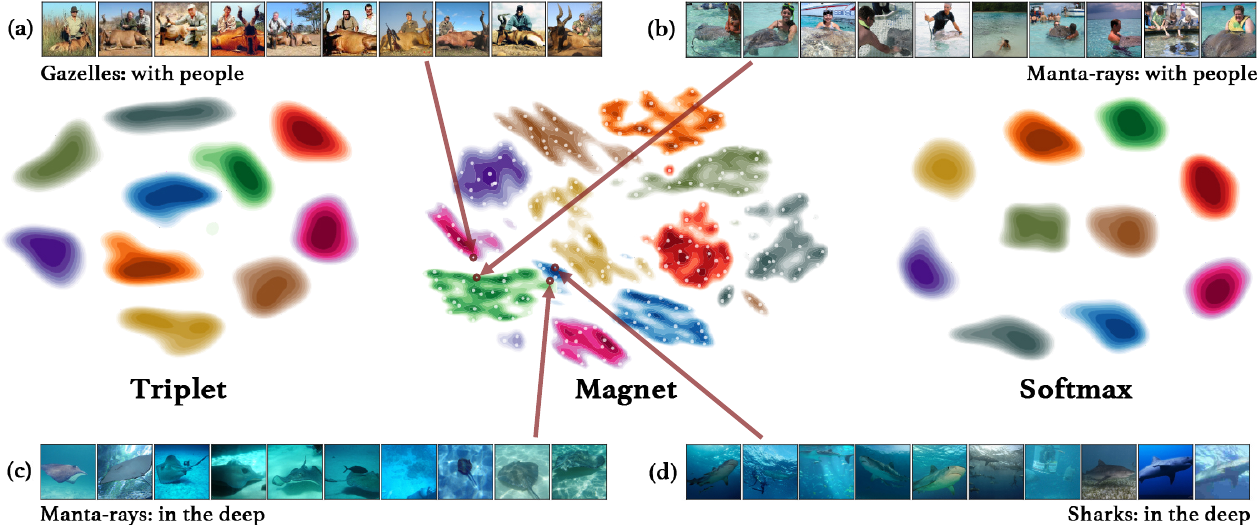}%
\caption{\small2D visualizations of representations attained by training triplet loss, \modelname{} and a softmax classifier on 10 classes of ImageNet. The {\bf different colours} correspond to different classes, and the values to density estimates computed from an application of t-SNE \citep{maaten2008visualizing} on the original 1024-dimensional representations. The {\bf white dots} in the \modelnameshort{} t-SNE correspond to $K=32$ clusters used by Magnet to capture each class. The {\bf red arrows} retrieve the examples closest to particular clusters (which were learnt autonomously). {\bf 1.} It can be seen that triplet loss and softmax result in unimodal separation, due to enforcement of semantic similarity. For \modelname{}, the distributions of the different classes may arbitrarily split, adaptively embracing intra-class variation and inter-class similarity. {\bf 2.} Green corresponds to manta-rays, blue to sharks, and magenta to gazelles. \modelname{} captures intra-class variation between (c) and (b) as manta-rays in the deep, and manta-rays with people. It also respects inter-class similarity, allowing shared structure between (c) and (d) as fish in the deep, and between (a) and (b) as animals with people. See Appendix \ref{app:tsne} for image maps of other t-SNE projections.}
\label{fig:tsne}
\end{minipage}
\end{widepage}
\end{figure*}

Instead, what we ought to do is rather define similarity as function of distances of our \emph{representations}~---~which lie in precisely the space sculpted for metric saliency. Since representations are adjusted continuously during training, it then follows that similarity must be defined adaptively. To that end, we must alternate between updating our representations, and refreshing our model which designates similarity as function of these. Visualizations of representations of different DML approaches can be found in a toy example in Figure \ref{fig:tsne}.

%%%%%%%%%%%%%%%%%%%%%%%%
\paragraph{Issue \#2: objective formulation}\label{formulation}
Two very popular classes of DML approaches have stemmed from Triplet Loss \citep{Weinberger:2009:DML:1577069.1577078} and Contrastive Loss \citep{Hadsell06dimensionalityreduction}. The outlined issues apply to both, but for simplicity of exposition we use triplet loss as an example. During its training, triplets consisting of a seed example, a ``positive'' example similar to the seed and a ``negative'' dissimilar example are sampled. Let us denote their representations as $\rmbr_m, \rmbr_m^+$ and $\rmbr_m^-$ for $m=1, \ldots, M$. Triplet loss then demands that the difference of distances of the representation of the seed to the negative and to the positive be larger than some pre-assigned margin constant $\alpha\in\reals$:
\begin{align}
\mathscr{L}_{\textrm{triplet}}\left(\bTheta\right) = \frac{1}{M}\sum_{m=1}^M \left\{
\left\|\rmbr_m - \rmbr^-_m\right\|^2_2
- \left\|\rmbr_m - \rmbr^+_m\right\|^2_2
+ \alpha
\right\}_+\;,
\end{align}
where $\left\{\cdot\right\}_+$ is the hinge function and $\bTheta$ the parameters of the map to representation space. The representations are often normalized to achieve scale invariance, and negative examples are mined in order to find margin violators (for example, \citet{Schroff_2015_CVPR,NIPS2012_4808}).

Objectives formulated in this spirit exhibit a short-sightedness. Namely, penalizing individual pairs or triplets of examples does not employ sufficient contextual insight of neighbourhood structure, and as such different triplet terms are not necessarily consistent. This hinders both the convergence rate as well as performance of these approaches. Moreover, the cubic growth of the number of triplets renders operation on these computationally inefficient.

In contrast to this, it is desirable to instead inform the algorithm of the \emph{distributions} of the different classes in representation space and their overlaps, and rather manipulate these in a way that is globally consistent. We elaborate on this in the section below.

\begin{figure*}[t]
\centering%
\subfigure[Triplet: before.]{
  \includegraphics[height=1.0in,trim=0cm 0cm 0cm 0cm,clip]{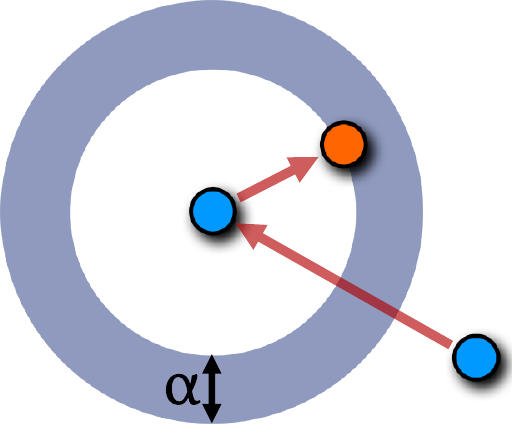}
  \label{fig:triplet_before}}
  \hfill
\subfigure[Triplet: after.]{
  \includegraphics[height=1.0in,trim=0cm 0cm 0cm 0cm,clip]{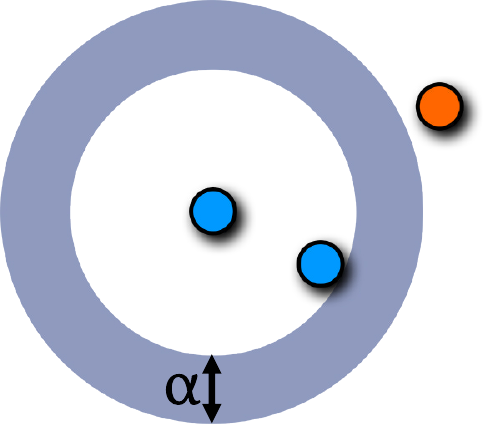}
  \label{fig:triplet_after}}
  \hfill
\subfigure[\modelnameshort{}: before.]{
  \includegraphics[height=1.0in,trim=0cm 0cm 0cm 0cm,clip]{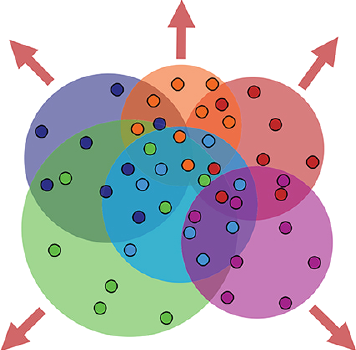}
  \label{fig:magnet_before}}
  \hfill
\subfigure[\modelnameshort{}: after.]{
  \includegraphics[height=1.0in,trim=0cm 0cm 0cm 0cm,clip]{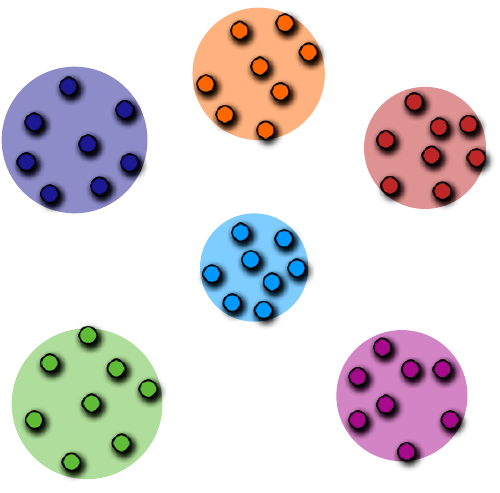}
  \label{fig:magnet_after}}
\caption{The intuition behind triplet loss and \modelname{}. Triplet loss only considers a single triplet at a time, resulting in reduced performance and training inefficiencies. In contrast, in \modelname{}, at each iteration an entire local neighbourhood of nearest clusters is retrieved, and their overlaps are penalized. Insight into representation distribution permits adaptive similarity characterization, local discrimination and a globally consistent optimization procedure.}
\label{fig:cartoon}
\end{figure*}

%%%%%%%%%%%%%%%%%%%%%%%%%%%%%%%%%%%%%%%%%%%%%%%%%%%%%%%%%%%%%%%%%%%%%%%%
\section{\modelname{} for Distance Metric Learning}\label{sec:magnet}
%%%%%%%%%%%%%%%%%%%%%%%%
We proceed to design a model to mitigate the identified difficulties. Let us for a moment neglect practical considerations, and envision our ideal DML approach. To start, as concluded at the start of Section \ref{variation}, we are interested to characterize similarity adaptively as function of current representation structure. We would then utilize this knowledge to pursue local separation as opposed to global: we seek to separate between distributions of different classes in representation space, but do not mind if they are interleaved. As such, let us assume that we have knowledge of the representation distribution of each class at any time during training. Our DML algorithm, then, would discover regions of local overlap between different classes, and penalize these to achieve discrimination. 

Such an approach would liberate us from the unimodality assumption and unreasonable prior target neighbourhood assignments --- resulting in a more expressive representation which maintains significantly more information. Moreover, employing a loss informed of distributions rather than individual examples would allow for a more coherent training procedure, where the distance metric is adjusted in a way that is globally consistent.

To that end, a natural approach would be to employ clustering techniques to capture these distributions in representation space. Namely, for each class, we will maintain an index of clusters, which we will update continuously throughout training. Our objective, then, would jointly manipulate entire clusters --- as opposed to individual examples --- in the pursuit of local discrimination. This intuition of cluster attraction and repulsion motivates us to name it \emph{\modelname{}}. A caricature illustrating the intuition behind this approach can be found in Figure \ref{fig:cartoon}.

In addition to its advantages from a modeling perspective, a clustering-based approach also facilitates computation by enabling efficient hard negative mining. That is, we may perform approximate nearest neighbour retrieval in a two-step process, where we first retrieve nearest clusters, after which we retrieve examples from these clusters. 

Finally, as discussed, throughout training we are interested in a more complete characterization of neighbourhood structure. At each iteration, we sample \emph{entire local neighbourhoods} rather than collections of independent examples (or triplets) as per usual, which significantly improves training efficiency. We elaborate on this in Section \ref{sec:training}.

%%%%%%%%%%%%%%%%%%%%%%%%
\subsection{Model formulation}\label{model}
We proceed to quantify the modeling objectives outlined above. Let us assume we have a training set consisting of $N$ input-label pairs $\mcD = \{\rmbx_n, y_n\}_{n=1}^N$ belonging to $C$ classes. We consider a parametrized map $\rmbf(\cdot; \bTheta)$ which hashes our inputs to \emph{representation space}, and denote their representations as $\rmbr_n = \rmbf(\rmbx_n; \bTheta), n = 1,\ldots, N$. In this work, we select this transformation as GoogLeNet \citep{43022,DBLP:conf/icml/IoffeS15}, which has been demonstrated to be a powerful CNN architecture; in Section \ref{sec:experiments} we elaborate on this choice.

We assume that, for each class $c$, we have $K$ cluster assignments $\mcI^c_1, \ldots, \mcI^c_{K}$ obtained via an application of the K-means algorithm. Note that $K$ may vary across classes, but for simplicity of exposition we fix it as uniform. In Section \ref{clustering_index}, we discuss how to maintain this index. To that end, we assume that these assignments have been chosen to minimize intra-cluster distances. Namely, for each class $c$, we have
\begin{eqnarray}
\mcI^c_1, \ldots, \mcI^c_K &=& \arg\min_{I^c_1, \ldots, I^c_K} \sum_{k=1}^K \sum_{\rmbr\in I^c_k} \left\|\rmbr - \bmu^c_k\right\|^2_2\;, \\
\bmu^c_k &=& \frac{1}{|I^c_k|}\sum_{\rmbr\in I^c_k} \rmbr\;.
\end{eqnarray}
We further define $C(\rmbr)$ as the class of representation $\rmbr$, and $\bmu(\rmbr)$ as its assigned cluster center.

We proceed to define our objective as follows:
\begin{align}
\mathscr{L}\left(\bTheta\right) = \frac{1}{N} \sum_{n=1}^N \left\{-\log
\frac{e^{-\frac{1}{2\sigma^2} \left\|\rmbr_n - \bmu(\rmbr_n)\right\|^2_2 - \alpha}}
{\sum_{c\neq C(\rmbr_n)}\sum_{k=1}^K e^{-\frac{1}{2\sigma^2} \left\|\rmbr_n - \bmu^c_k\right\|^2_2}}
\right\}_+ \label{eqn:objective}
\end{align}
where $\left\{\cdot\right\}_+$ is the hinge function, $\alpha\in\reals$ is a scalar, and $\sigma^2=\frac{1}{N-1} \sum_{\rmbr\in \mcD} \left\|\rmbr - \bmu(\rmbr)\right\|^2_2$ is the variance of all examples away from their respective centers. We note that cluster centers sufficiently far from a particular example vanish from its term in the objective. This allows accurately approximating each term with a small number of nearest clusters. 

A feature of this objective not usually available in standard distance metric learning approach is variance standardization. This renders the objective invariant to the characteristic lengthscale of the problem, and allows the model to gauge its confidence of prediction by comparison of intra- and inter-cluster distances. With this in mind, $\alpha$ is then the desired cluster separation gap, measured in units of variance. In our formulation, we may thus interpret $\alpha$ as a modulator of the probability assigned to an example of a particular class under the distribution of another.

We remark that during model design, an alternative objective we considered is the cluster-based analogue of NCA (see Section \ref{relations}): this objective seems to be a natural approach with a clear probabilistic interpretation. However, we found empirically that this objective does not generalize as well, since it only vanishes in the limit of extreme discrimination margins. 

%%%%%%%%%%%%%%%%%%%%%%%%
\subsection{Training procedure}\label{sec:training}
\paragraph{Component \#1: neighbourhood sampling} At each iteration, we sample entire local neighbourhoods rather than a collection of independent examples. Namely, we construct our minibatch in the following way:
\begin{enumerate}
  \item Sample a seed cluster $I_1\sim p_{\mcI}(\cdot)$
  \item Retrieve $M-1$ nearest impostor clusters $I_2, \ldots, I_M$ of $I_1$
  \item For each cluster $I_m, m = 1,\ldots, M$, sample $D$ examples $\rmbx^m_1,\ldots\rmbx^m_D\sim p_{I_m}(\cdot)$
\end{enumerate}
The choices of $p_{\mcI}(\cdot)$ and $p_{I_m}(\cdot), m = 1, \ldots, M$ allow us to adapt to the current distributions of examples in representation space. Namely, in our training, these allow us to specifically target and reprimand contested neighbourhoods with large cluster overlap. During training, we cache the losses of individual examples, from which we compute the mean loss $\mathscr{L}_I$ of each cluster $I$. We then choose $p_{\mcI}(I)\propto\mathscr{L}_I$, and $P_{I_m}(\cdot)$ as a uniform distribution. We remark that these choices work well in practice, but have been made arbitrarily and perhaps can be improved.

Given our samples, we may proceed to construct a stochastic approximation of our objective:
\begin{align}
\hat{\mathscr{L}}\left(\bTheta\right) = \frac{1}{MD} \sum_{m=1}^M \sum_{d=1}^D \left\{-\log
\frac{e^{-\frac{1}{2\hsigma^2} \left\|\rmbr^m_d - \bhmu_m\right\|^2_2 - \alpha}}
{\sum_{\bhmu\,:\,C(\bhmu)\neq C(\rmbr^m_d)} e^{-\frac{1}{2\hsigma^2} \left\|\rmbr^m_d - \bhmu\right\|^2_2}}
\right\}_+
\end{align}
where we approximate the cluster means as $\bhmu_m = \frac{1}{D}\sum_{d=1}^D \rmbr^m_d$ and variance as ${\hsigma=\frac{1}{MD-1} \sum_{m=1}^M \sum_{d=1}^D \left\| \rmbr^m_d - \bhmu_m\right\|^2_2}$. During training, we backpropagate through this objective, and the full CNN which gave rise to the representations.

\paragraph{Component \#2: cluster index}\label{clustering_index}
As mentioned above, we maintain for each class a K-means index which captures its distribution in representation space during training. We initialize each index with K-means++ \citep{Arthur:2007:KAC:1283383.1283494}, and refresh it periodically. To attain the representations to be indexed, we pause training and compute the forward passes of all inputs in the training set. The computational cost of refreshing the cluster index is significantly smaller than the cost of training the CNN itself: it is not done frequently, it only requires forward passes of the inputs, and the relative cost of K-means clustering is negligible. 

It may seem that freezing the training is unnecessarily computationally expensive. Note that we also explored the alternative strategy of caching the representations of each minibatch on-the-fly during training. However, we found that it is critical to maintain the true neighbourhood structure where the representations are all computed in the same stage of learning. We empirically observed that since the representation space is changing continuously during training, indexing examples whose representations were computed in different times resulted in incorrect inference of neighbourhood structure, which in turn irreparably damaged nearest impostor assessment.

\paragraph{Improvement of training efficiency} The proposed approach offers a number of benefits which compound to considerably enhance training efficiency, as can be seen empirically in Section \ref{sec:fine_grained}. First, one of the main criticisms of triplet-based approaches is the cubic growth of the number of triplets. Manipulating entire clusters of examples, on the other hand, significantly improves this complexity, as this requires far fewer pairwise distance evaluations. Second, operating on entire cluster neighbourhoods also permits information recycling: we may jointly separate all clusters from one another at once, whereas an approach based on independent sampling would require far more repetitions of the same examples. Finally, penalizing clusters of points away from one another leads to a more coherent adjustment of each point, whereas different triplet terms may not necessarily be consistent with one another. 

%%%%%%%%%%%%%%%%%%%%%%%%
\subsection{Evaluation procedure}\label{evaluation_procedure}
The evaluation procedure is consistent with the objective formulation: we assign the label of each example $\rmbx_n$ as function of its representation's softmax similarities to its $L$ closest clusters, say $\bmu_1, \ldots, \bmu_L$. More precisely, we choose label $c^*_n$ as
\begin{align}
 c^*_n = \arg\max_{c=1, \ldots, C}
\frac{\sum_{\bmu_l\,:\,C(\bmu_l)=c} e^{-\frac{1}{2\sigma^2} \left\|\rmbr_n - \bmu_l\right\|^2_2}}
{\sum_{l=1}^L e^{-\frac{1}{2\sigma^2} \left\|\rmbr_n - \bmu_l\right\|^2_2}}\;,
\label{eqn:evaluation}
\end{align}
where $\sigma$ is a running average of stochastic estimates $\hat{\sigma}$ computed during training.

This can be thought of as ``$k$-nearest-cluster'' (kNC), a variant of a soft kNN classifier. This has the added benefit of reducing the complexity of nearest neighbour evaluation from being a function of the number of examples to the number of clusters. Here, the lengthscale $\sigma$ autonomously characterizes local neighbourhood radius, and as such implies how to sensibly choose $L$. In general, we found that performance improves monotonically with $L$, as the soft classification is able to make use of additional neighbourhood information. At some point, however, retrieving additional nearest neighbours is clearly of no further utility, since these are much farther away than the lengthscale defined by $\sigma$. In practice we use $L=128$ for all experiments in this work.

%%%%%%%%%%%%%%%%%%%%%%%%
\subsection{Relation to existing models}\label{relations}
\paragraph{Triplet Loss} Our objective proposed in Equation \ref{eqn:objective} has the nice property that it reduces to the familiar triplet loss under a particular set of assumptions. Specifically, let us assume that we approximate each neighbourhood with a \emph{single} impostor cluster, i.e, $M=2$. Let us further assume that we approximate the seed cluster with merely $D=2$ samples, and the impostor cluster with one. We further simplify by ignoring the variance normalization. Our objective then exactly reduces to triplet loss for a pair of triplets ``symmetrized'' for the two positive examples:
\begin{align}
\hat{\mathscr{L}}\left(\bTheta\right) = \sum_{d=1}^2 \left\{
\left\|\rmbr^1_d - \rmbr^1_{2-d}\right\|^2_2
-\left\|\rmbr^1_d - \rmbr^2_1\right\|^2_2
+ \alpha
\right\}_+\;.
\end{align}

\paragraph{Neighbourhood Components Analysis} Neighbourhood Components Analysis (NCA) and its extensions \citep{Goldberger04neighbourhoodcomponents,AISTATS07_SalakhutdinovH,DBLP:conf/icml/MinMYBZ10} have been designed in a similar spirit to \modelname{}. The NCA objective is given by
\begin{align}
\mathscr{L}_{\textrm{NCA}}\left(\bTheta\right) = \frac{1}{N} \sum_{n=1}^N -\log
\frac{\sum_{n'\,:\,C(\rmbr_{n'}) = C(\rmbr_{n})} e^{-\left\|\rmbr_n - \rmbr_n'\right\|^2_2}}
{\sum_{n' = 1}^N e^{-\left\|\rmbr_n - \rmbr_n'\right\|^2_2}}\;.
\label{eqn:nca}
\end{align}
However, this formulation does not address a number of concerns both in modeling and implementation. As an example, it does not touch on minibatch sampling in large datasets. Even if we maintain a nearest neighbour index, if we na\"{i}vely retrieve the nearest neighbours for each example, they are all going to be of different classes with high probability. 

\paragraph{Nearest Class Mean} Nearest Class Mean \citep{mensink13pami} is cleverly designed for scalable DML. In this approach, the mean vectors $\bmu_c=\frac{1}{|C|} \sum_{c(\rmbx)=c} \rmbx, c=1,\ldots, C$ of the examples in their raw input form are computed and fixed for each class. A linear transformation $\rmbW$ is then learned to maximize the softmax distance of each example to the cluster center of its class:
\begin{align}
\mathscr{L}_{\textrm{NCM}}\left(\rmbW\right) = \frac{1}{N} \sum_{n=1}^N -\log
\frac{e^{-\left\|\rmbW\rmbx_n - \rmbW\bmu_{c(\rmbx_n)}\right\|^2_2}}
{\sum_{c = 1}^C e^{-\left\|\rmbW\rmbx_n - \rmbW\bmu_c\right\|^2_2}}\;.
\label{eqn:ncm}
\end{align}

The authors further generalize this to Nearest Class Multiple Centroids (NCMC), where for each class, $K$ centroids are computed with K-means. Magnet shares many ideas with NCMC, but these approaches differ in a number of important ways. For NCMC, the centroids are computed on the raw inputs and are fixed prior to training, rather than updated continuously on a learnt representation. It is also not clear how to extend this to more expressive transformations (such as CNNs) to representation space, but this step is required in order to enjoy the success of deep learning approaches in a DML setting.

%%%%%%%%%%%%%%%%%%%%%%%%%%%%%%%%
\begin{figure*}[t]

\begin{widepage}
\centering%
\hfill\subfigure[Stanford Dogs.]{
    \begin{tabular}[t]{lr}
    \toprule
    {\bf Approach}          & {\bf Error}\\
    \cmidrule(r){1-2}%
    \small\citeauthor{DBLP:conf/wacv/AngelovaL14} & \small51.7\%\\
    \small\citeauthor{GavvesICCV2013}   & \small49.9\%\\
    \small\citeauthor{DBLP:conf/cvpr/XieYWL15}  & \small43.0\%\\
    \small\citeauthor{gavves2015}  & \small43.0\%\\
    \small\citeauthor{Qian_2015_CVPR}   & \small30.9\%\\
   \cmidrule(r){1-2}
    \small Softmax                 & \small26.6\%\\      
    \small Triplet                 & \small35.8\% \\    
    \small{\bf \modelnameshort{}}            & {\bf 2\small4.9\%} \\
   \bottomrule%
   \newline

\end{tabular}}
  \hfill
\subfigure[Oxford 102 Flowers.]{
\begin{tabular}[t]{lr}
    \toprule
    {\bf Approach}          & {\bf Error}\\
    \cmidrule(r){1-2}%
    \small\citeauthor{DBLP:conf/cvpr/AngelovaZ13} & \small23.3\%\\
    \small\citeauthor{DBLP:conf/wacv/AngelovaL14} & \small19.6\%\\
    \small\citeauthor{DBLP:conf/cvpr/MurrayP14} & \small15.4\%\\
    \small\citeauthor{Razavian_2014_CVPR_Workshops} & \small13.2\%\\
    \small\citeauthor{Qian_2015_CVPR} & \small11.6\%\\
   \cmidrule(r){1-2}
    \small Softmax                 & \small11.2\%\\      
    \small Triplet                 & \small17.0\%\\    
    \small{\bf \modelnameshort{}}            & {\bf\small 8.6\%}\\
   \bottomrule%
   \newline
\end{tabular}}
\hfill
\subfigure[Oxford-IIIT Pet.]{
  \begin{tabular}[t]{lr}
    \toprule
    {\bf Approach}          & {\bf Error}\\
    \cmidrule(r){1-2}%
    \small\citeauthor{DBLP:conf/cvpr/AngelovaZ13} & \small49.2\%\\
    \small\citeauthor{parkhi12a} & \small46.0\%\\
    \small\citeauthor{DBLP:conf/wacv/AngelovaL14} & \small44.6\%\\
    \small\citeauthor{DBLP:conf/cvpr/MurrayP14} & \small43.2\%\\
    \small\citeauthor{Qian_2015_CVPR} & \small19.6\%\\
   \cmidrule(r){1-2}
    \small Softmax                 & \small11.3\%\\      
    \small Triplet                 & \small13.5\%\\    
    \small{\bf \modelnameshort{}}            & {\bf \small10.6\%}\\
   \bottomrule%
   \newline
\end{tabular}}\hspace{\fill}

\hfill\quad\;\subfigure[ImageNet Attributes.]{
        \begin{tabular}[t]{lr}
        \toprule
        {\bf Approach}          & {\bf Error}\\
       \cmidrule(r){1-2}
        \small{\bf Softmax}                 & \small{\bf 14.1\%}\\      
        \small Triplet                 & \small26.8\% \\    
        \small\modelnameshort{}                  & \small15.9\% \\
       \bottomrule%
       \newline
    \end{tabular}
}
\quad
\subfigure[Different metrics on Stanford Dogs.]{
    \raisebox{-1.0\height}{\includegraphics[height=0.9in,trim=0cm 0cm 0cm 0cm,clip]{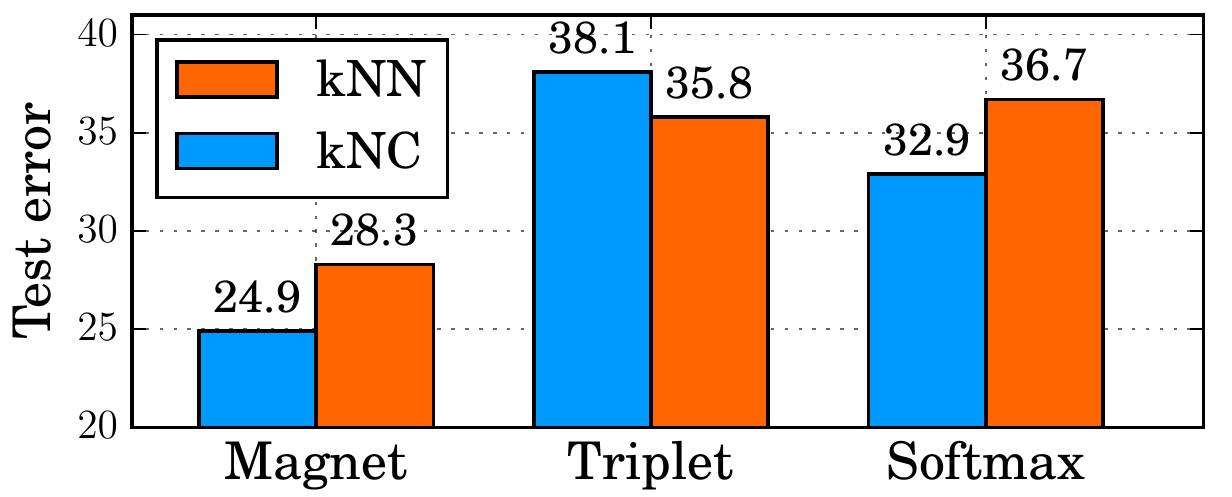}}
\label{fig:metrics}}
\subfigure[Hierarchy recovery on ImageNet Attributes.]{
    \raisebox{-1.0\height}{
    \begin{minipage}{2.4in}
    \centering
        \begin{tabular}[t]{lrr}
        \toprule
        {\bf Approach}          & \small{\bf Error@1} & \small{\bf Error@5}\\
       \cmidrule(r){1-3}
        \small Softmax                 & \small30.9\% & \small15.0\%\\      
        \small Triplet                 & \small44.6\% & \small23.4\%\\    
        \small{\bf \modelnameshort{}} & \small{\bf 28.6\%} & \small{\bf 7.8\%}\\
       \bottomrule%
       \newline
    \end{tabular}
    \end{minipage}}
\label{fig:hierarchy}}\hspace{\fill}
\caption{{\bf(a)-(d)} Comparison of test set errors of various state-of-the-art approaches on different fine-grained visual categorization datasets. The bottom three results for each table were all attained by applying different objectives on exactly the same architecture. {\bf (e)} Evaluation of test errors on the Stanford Dogs dataset under different metrics. {\bf (f)} We explore whether each algorithm is able to recover a latent class hierarchy, provided only coarse superclasses. We collapse random pairs of classes of ImageNet Attributes onto the same label. We then train on the corrupted labels, and report test errors on the original classes.}
\label{fig:finegrained}
\end{widepage}
\end{figure*}

%%%%%%%%%%%%%%%%%%%%%%%%%%%%%%%%%%%%%%%%%%%%%%%%%%%%%%%%%%%%%%%%%%%%%%%%
\section{Experiments}\label{sec:experiments}
We run all experiments on a cluster of Tesla K40M GPU's. All parametrized maps $\rmbf(\cdot; \bTheta)$ to representation space are chosen as GoogLeNet with batch normalization \citep{DBLP:conf/icml/IoffeS15}. We add an additional fully-connected layer to map to a representation space of dimension 1024. 

We find that it is useful to warm-start any DML optimization with weights of a partly-trained a standard softmax classifier. It is important to not use weights of a net trained to completion, as this would result in information dissipation and as such defeat the purpose of pursuing DML in the first place. Hence, we initialize all models with the weights of a net trained on ImageNet \citep{ILSVRC15} for 3 epochs only. We augment all experiments with random input rescaling of up to 30\%, followed by jittering back to the original input size of $224\times 224$. At test-time we evaluate an input by averaging the outputs of 16 random samples drawn from this augmentation distribution. 

%%%%%%%%%%%%%%%%%%%%%%%%
\subsection{Fine-grained classification}\label{sec:fine_grained}
We validate the classification efficacy of the learnt representations on a number of popular fine-grained visual categorization tasks, including Stanford Dogs \citep{KhoslaYaoJayadevaprakashFeiFei_FGVC2011}, Oxford-IIIT Pet \citep{parkhi12a} and Oxford 102 Flowers \citep{Nilsback08} datasets. We also include results on ImageNet attributes, a dataset described in Section \ref{sec:attributes}. 

We seek to compare optimal performances of the different \emph{model spaces}, and so perform hyperparameter search on validation error generated by 3 classes of objectives: a standard softmax classifier, triplet loss, and \modelname{}. The hyperparameter search setup, including optimal configurations for each experiment, is specified in full detail in Appendix \ref{app:optimal}. In general, for \modelname{} we observed empirically that it is beneficial to increase the number of clusters per minibatch to around $M=12$ in the cost of reducing the number of retrieved examples per cluster to $D=4$. The optimal gap has in general been $\alpha\approx 1$, and the value of $K$ varied as function of dataset cardinality. 

The classification results can be found in Table \ref{fig:finegrained}. We use soft kNN to evaluate triplet loss error and kNC (see Section \ref{evaluation_procedure}) for \modelname{}. However, for completeness of comparison, in Figure \ref{fig:metrics} we present evaluations of all learnt representations under both kNN and kNC. 

It can be observed that \modelname{} outperforms the traditional triplet loss by a considerable margin. It is also able to surpass the standard softmax classifier in most cases: while the margin is not significant, note that the true win here is in terms of learning representations much more suitable for task transfer, as validated in the following subsections.

\begin{figure*}[t]
\centering%
\includegraphics[width=\textwidth,trim=0cm 0cm 0cm 0cm,clip]{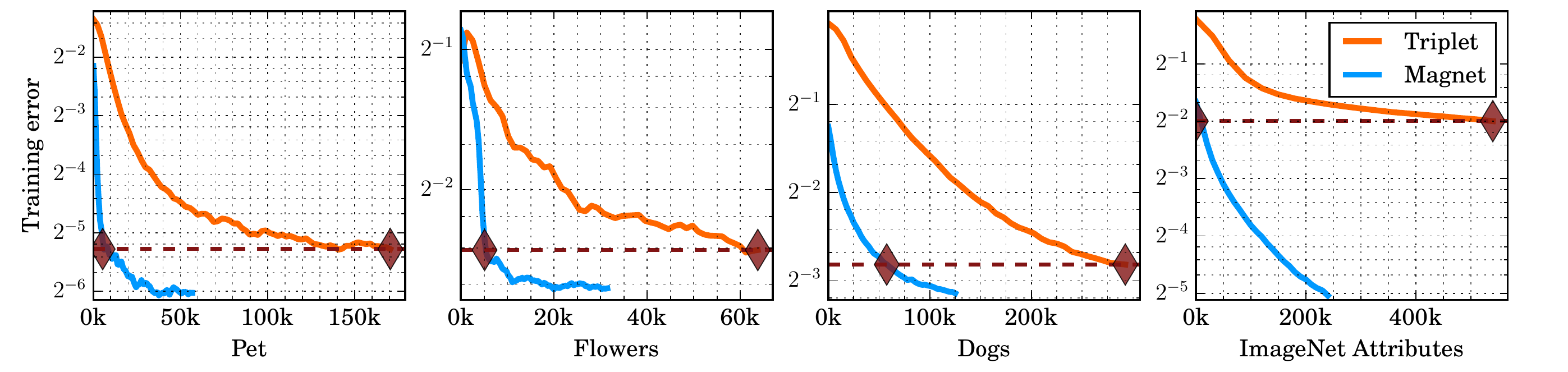}%
\caption{Training curves for various experiments as function of number of iterations. For both triplet and \modelname{} objectives, the experiment with optimal hyperparameter configuration for each model space is presented. The red diamonds indicate the point in time in which the triplet asymptotic error rate is achieved. It can be observed that \modelname{} reaches the same error in 5-30 times fewer iterations.}
\label{fig:learning_dynamics}
\end{figure*}

In Figure \ref{fig:learning_dynamics}, it can be seen that \modelname{} reaches the triplet loss asymptotic error rate 5-30 times faster. The prohibitively slow convergence of triplet loss has been well-known in the community. \modelname{} achieves this speedup as it mitigates some of the training-time inefficiencies featured by triplet loss presented throughout Section \ref{challenges} and the end of Section \ref{sec:training}. For fairness of comparison, we remark that softmax converges faster than \modelnameshort{}; however, this comes at the cost of a less informative representation.

\subsection{Attribute distribution}\label{sec:attributes}
We expect \modelnameshort{} to sculpt a more expressive representation, which enables similar examples of different classes to be close together, and dissimilar examples of the same class to be far apart; this can be seen qualitatively in Figure \ref{fig:tsne}. In order to explore this hypothesis quantitatively, after training is complete we examine the attributes of neighbouring examples as a proxy for assessment of similarity. We indeed find the distributions of these attributes to be more concentrated for \modelnameshort{}.

\begin{figure*}[t]
\begin{widepage}
\centering%
\hfill
\subfigure[Attribute examples.]{
\begin{minipage}{1.1in}
\rotatebox{90}{\quad\ \ Furry}\hspace{0.01cm}
  \includegraphics[width=0.7in,trim=0cm 0cm 0cm 0cm,clip]{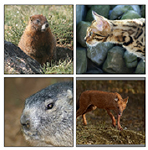}\vspace{0.2in}\\
  \rotatebox{90}{\quad Spotted}\hspace{0.04cm}
  \includegraphics[width=0.7in,trim=0cm 0cm 0cm 0cm,clip]{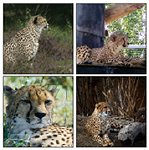}%
\end{minipage}
  \label{fig:attr_examples}}
  \quad
\subfigure[\modelnameshort{}.]{
\begin{minipage}{1.0in}
  \includegraphics[width=1.0in,trim=0cm 0cm 0cm 0cm,clip]{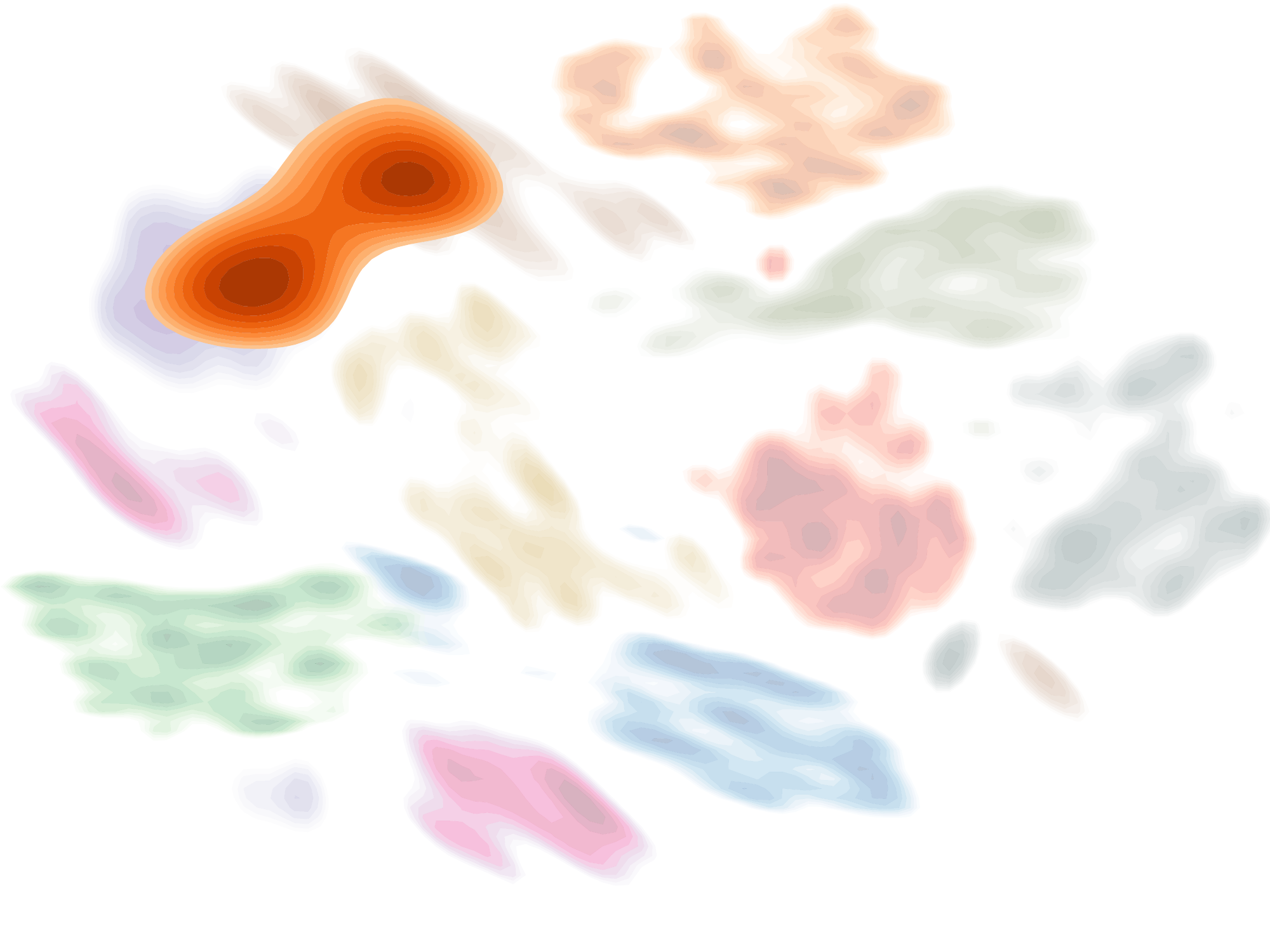}\vspace{0.1in}
  \includegraphics[width=1.0in,trim=0cm 0cm 0cm 0cm,clip]{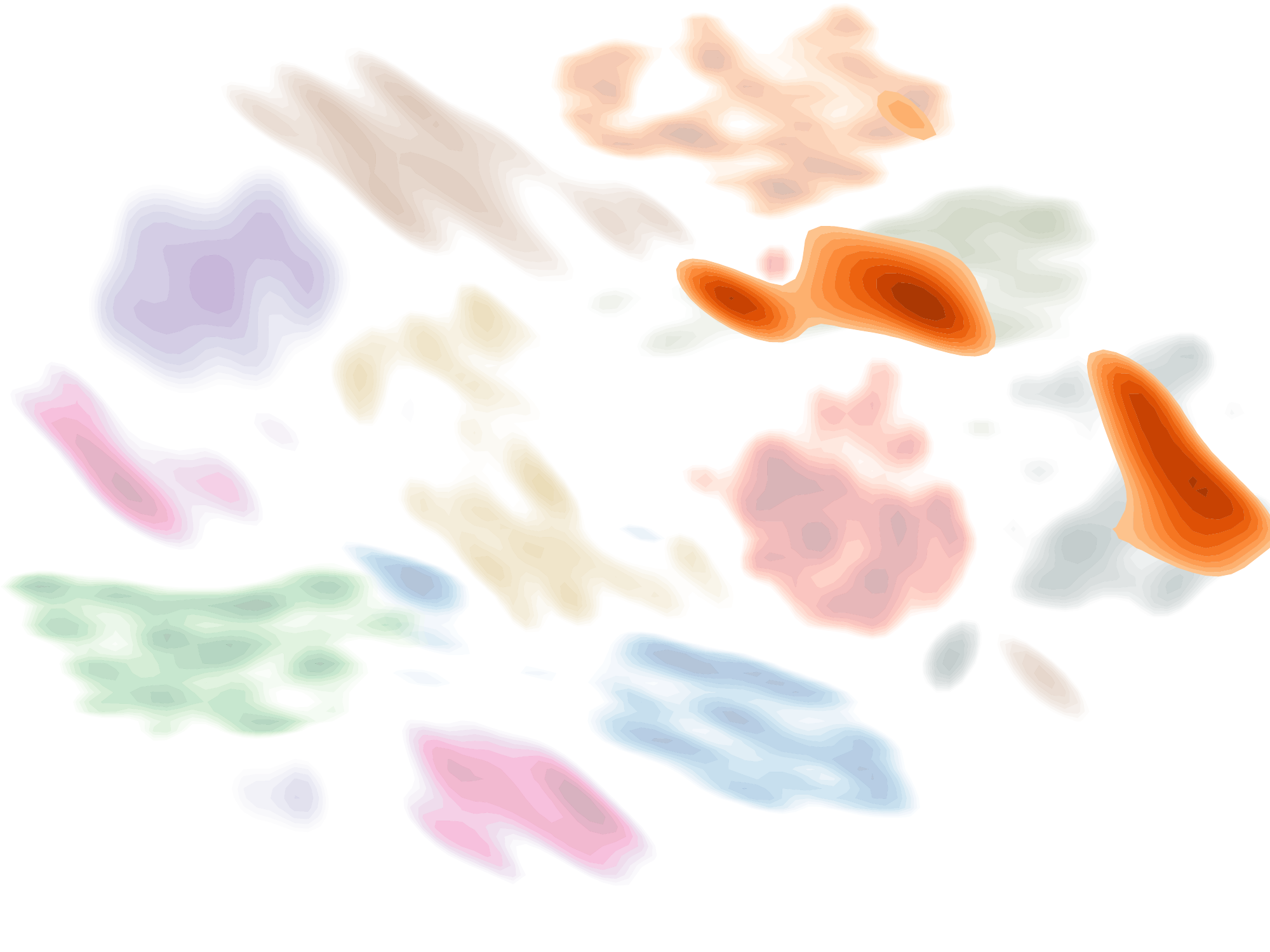}%
\end{minipage}
  \label{fig:attr_magnet}}
  \quad
\subfigure[Softmax.]{
\begin{minipage}{1.0in}
  \includegraphics[width=1.0in,trim=0cm 0cm 0cm 0cm,clip]{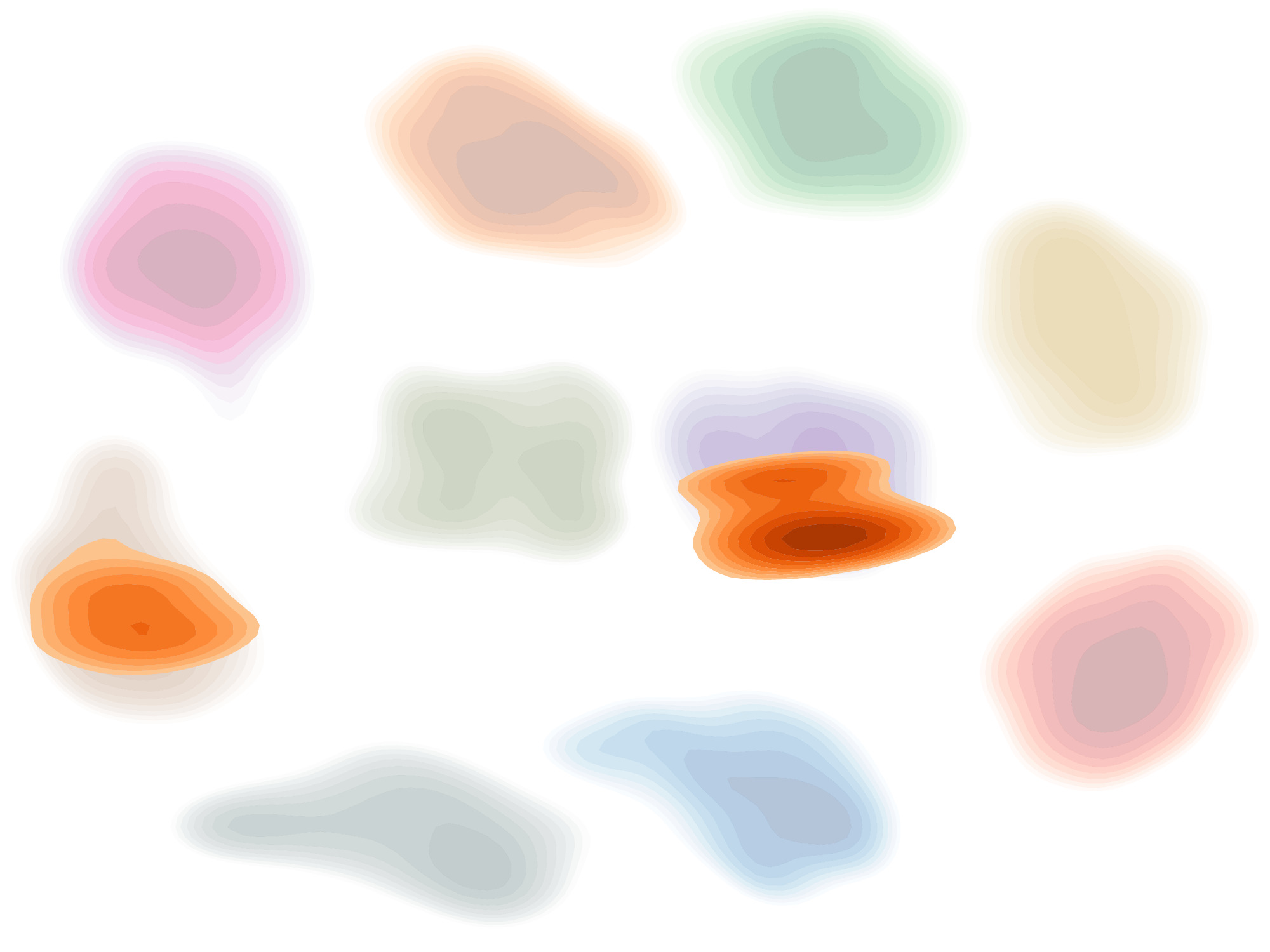}\vspace{0.1in}
  \includegraphics[width=1.0in,trim=0cm 0cm 0cm 0cm,clip]{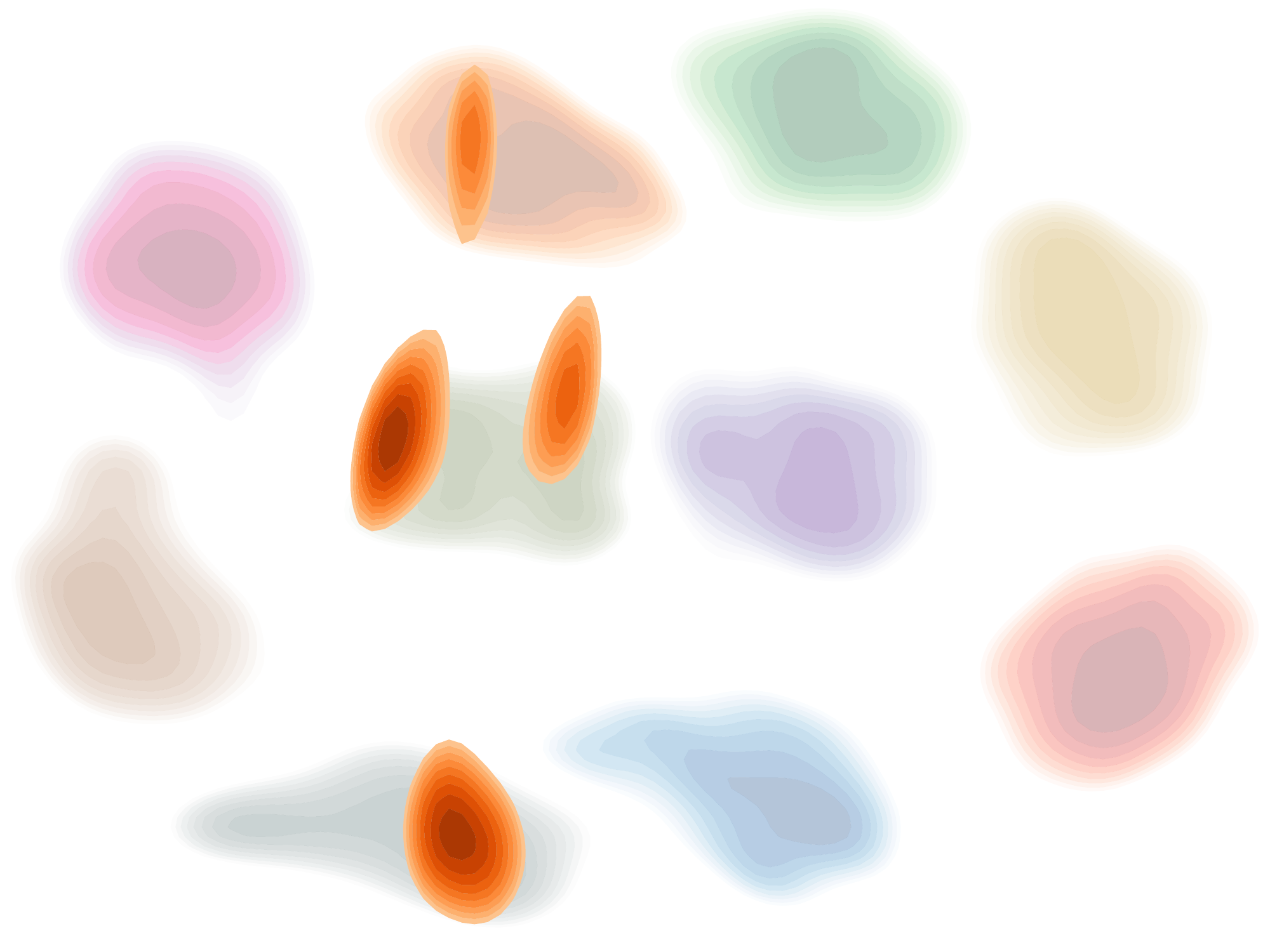}%
\end{minipage}
  \label{fig:attr_softmax}}
  \quad
\subfigure[Attribute precision.]{
\begin{minipage}{1.7in}
  \includegraphics[width=1.7in,trim=0cm 0cm 0cm 0cm,clip]{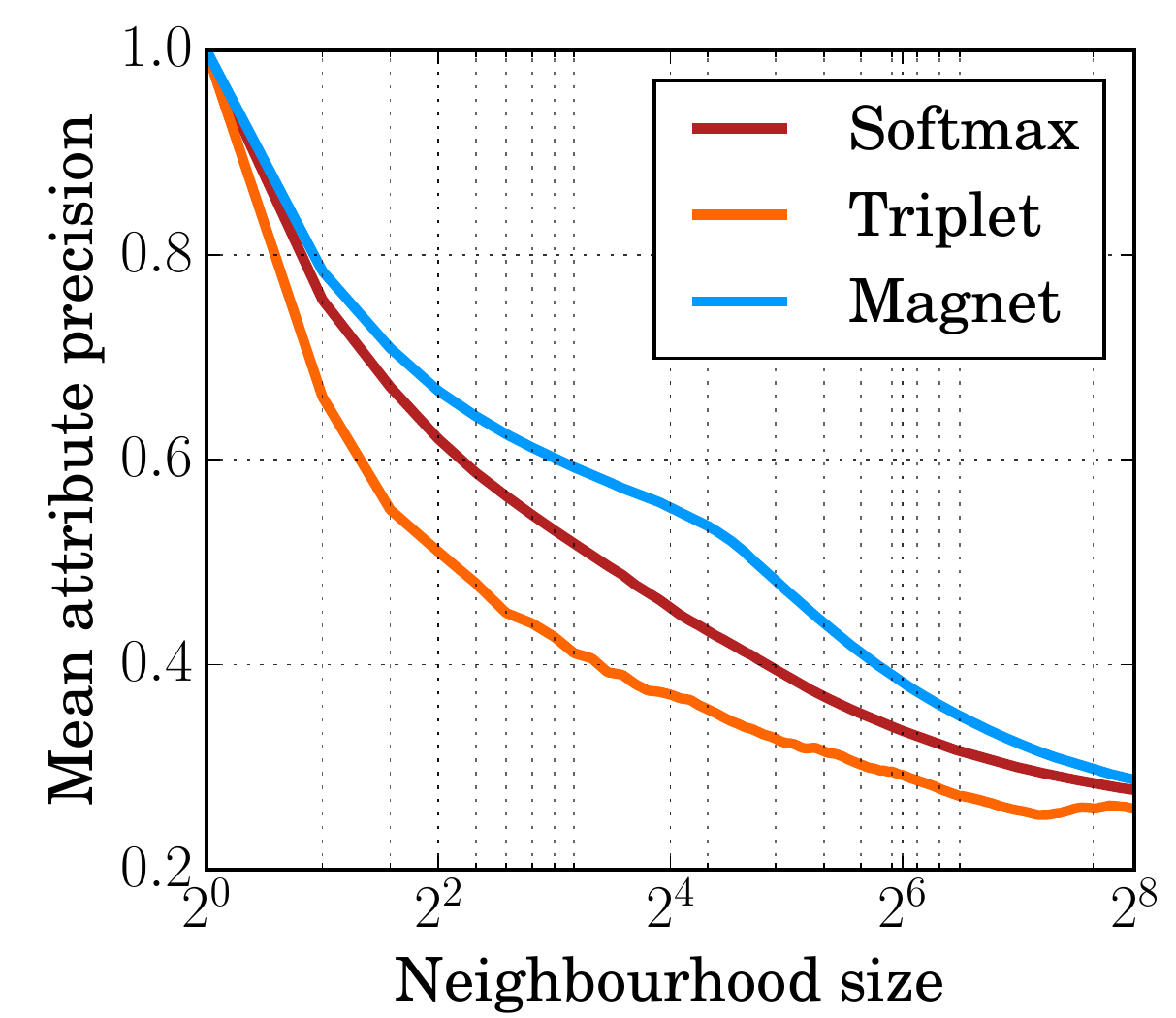}\vspace{0.1in}
\end{minipage}
  \label{fig:precision}}
  \hspace{\fill}
\caption{Attribute concentration properties. {\bf (a)} Examples of images featuring particular attributes. {\bf (b) \& (c)} The {\bf translucent underlying densities} correspond to the t-SNE visualizations presented in Figure \ref{fig:tsne}. These are overlaid with distributions of examples featuring the specified attributes, coloured in {\bf orange}. Magnet clusters together examples of different classes but with similar attributes, whereas softmax and triplet loss (not shown) do not. {\bf (d)} Mean fraction of neighbours featuring the same attributes as function of neighbourhood cardinality. Magnet consistently outperforms softmax and triplet across neighbourhood sizes.}
\end{widepage}
\label{fig:attributes}
\end{figure*}

We attain attribute labels from the Object Attributes dataset \citep{RussakovskyECCV10}. This provides 25 attribute annotations for 90 classes of an updated version of ImageNet, with about 25 annotated examples per class. Attributes include visual properties such as ``striped'', ``brown'', ``vegetation'' and so on; examples of these can be found in Figure \ref{fig:attr_examples}. Annotations are assigned individually for each input, which allows capturing intra-class variation and inter-class invariance.

We train softmax, triplet and \modelname{} objectives on a curated dataset we refer to as \emph{ImageNet Attributes}. This dataset contains 116,236 examples, and comprises all examples of each of the 90 ImageNet classes for which any attribute annotations are available: in Appendix \ref{app:im_attr} we describe it in detail. We emphasize we do not employ any attribute information during training. At convergence, we measure attribute concentration by computing mean attribute precision as function of neighbourhood size. Specifically, for each example and attribute, we compute over different neighbourhood cardinalities the fraction of neighbours also featuring this attribute. 

This result can be found in Figure \ref{fig:precision}. \modelname{} outperforms both softmax and triplet losses by a reasonable margin in terms of attribute concentration, with consistent gains of 25-50\% over triplet and 10-25\% over softmax across neighbourhood sizes. It may seem surprising that softmax surpasses triplet --- an approach specifically crafted for distance metric learning. However, note that while the softmax classifier requires high relative projection onto the hyperplane associated with each class, it leaves some flexibility for information retainment in its high-dimensional nullspace. Triplet loss, on the other hand, demands separation based on an imprecise assessment of similarity, resulting in poor proximity of similar examples of different classes.

\modelnameshort{}'s attribute concentration can also be observed visually in Figures \ref{fig:attr_magnet} and \ref{fig:attr_softmax}, presenting the t-SNE projections from Figure \ref{fig:tsne} overlaid with attribute distribution. It can be seen qualitatively that the \modelnameshort{} attributes are concentrated in particular areas of space, irrespective of class.

%%%%%%%%%%%%%%%%%%%%%%%%%%%%%%%%%%%%%%%%%%%%%%%%%%%%

\subsection{Hierarchy recovery}\label{sec:hierarchy_recovery}
In this experiment, we are interested to see whether each algorithm is able to recover a latent class hierarchy, provided only coarse superclasses. To test this, we randomly pair all classes of ImageNet Attributes, and collapse each pair under a single label. We then train on the corrupted labels, and check whether the finer-grained class labels may be recovered from the learnt representations. 

The results can be found in Table \ref{fig:hierarchy}. Magnet is able to identify intra-class representation variation, an essential property for success in this task. Softmax also achieves surprisingly competitive results, suggesting that meaningful variation is nevertheless captured within the nullspace of its last layer. For triplet loss, on the other hand, target neighbourhoods are designated prior to training, and as such it is not able to adaptively discriminate finer structure within superclasses.  

\section{Discussion and Future Work}
In this work, we highlighted a number of difficulties in a class of DML algorithms, and sought to address them. We validated the effectiveness of our approach under a variety of metrics, ranging from classification performance to convergence rate to attribute concentration.

In this paper, we anchored in place a number of parameters: we chose the number of clusters $K$ per class as uniform across classes, and refreshed our representation index at a fixed rate. We believe that adaptively varying these during training can enhance performance and facilitate computation. 

Another interesting line of work would be to replace the density estimation and indexing component with an approach more sophisticated than K-means. One natural candidate would be a tree-based algorithm. This would enable more efficient and more accurate neighbourhood retrieval.

\subsubsection*{Acknowledgements} 
We are grateful to Laurens van der Maaten, Florent Perronnin, Rob Fergus, Yaniv Taigman and others at Facebook AI Research for meaningful discussions and input. We thank Soumith Chintala, Alexey Spiridonov, Kevin Lee and everyone else at the FAIR Engineering team for unbreaking things at light speed.
\newline\newline\newline

%%%%%%%%%%%%%%%%%%%%%%%%%%%%%%%%%%%%%%%%%%%%%%%%%%%%%%%%%%%%%%%%%%%%%%%%%%%%%%%%%%%%%%%%%%%%%%%%%%%%%%%%
% \newpage
\begin{appendices}
\section{t-SNE image maps for typical Magnet and triplet representation spaces}\label{app:tsne}

\begin{figure*}[h]
\begin{widepage}
\centering%
\caption{Visualization of t-SNE map for a typical Magnet representation. We highlight interesting observations of the distributions of the learnt representations splitting to repsect intra-class variance and inter-class similarity.}\label{fig:tsne_appendix}
  \includegraphics[width=\textwidth,trim=0cm 0cm 0cm 0cm,clip]{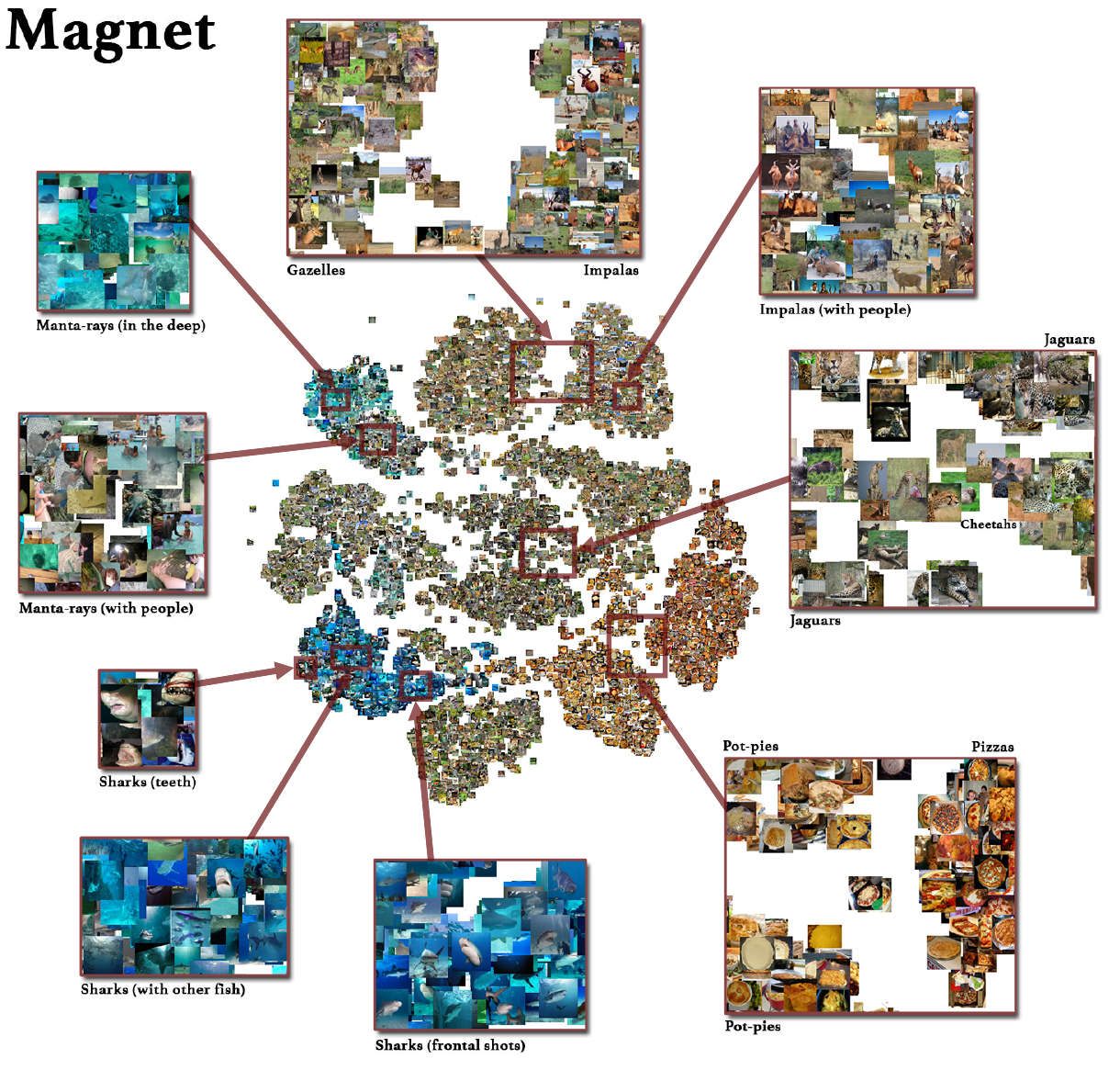}
\end{widepage}
\end{figure*}
\newpage
\begin{figure*}[h]
\begin{widepage}
\centering%
\caption{Visualization of t-SNE map for a typical triplet representation with enforcement of semantic similarity. Classes with similar examples are far from one another, and no obvious local similarity can be found within individual classes.}\label{fig:tsne_appendix}
  \includegraphics[width=0.85\textwidth,trim=0cm 0cm 0cm 0cm,clip]{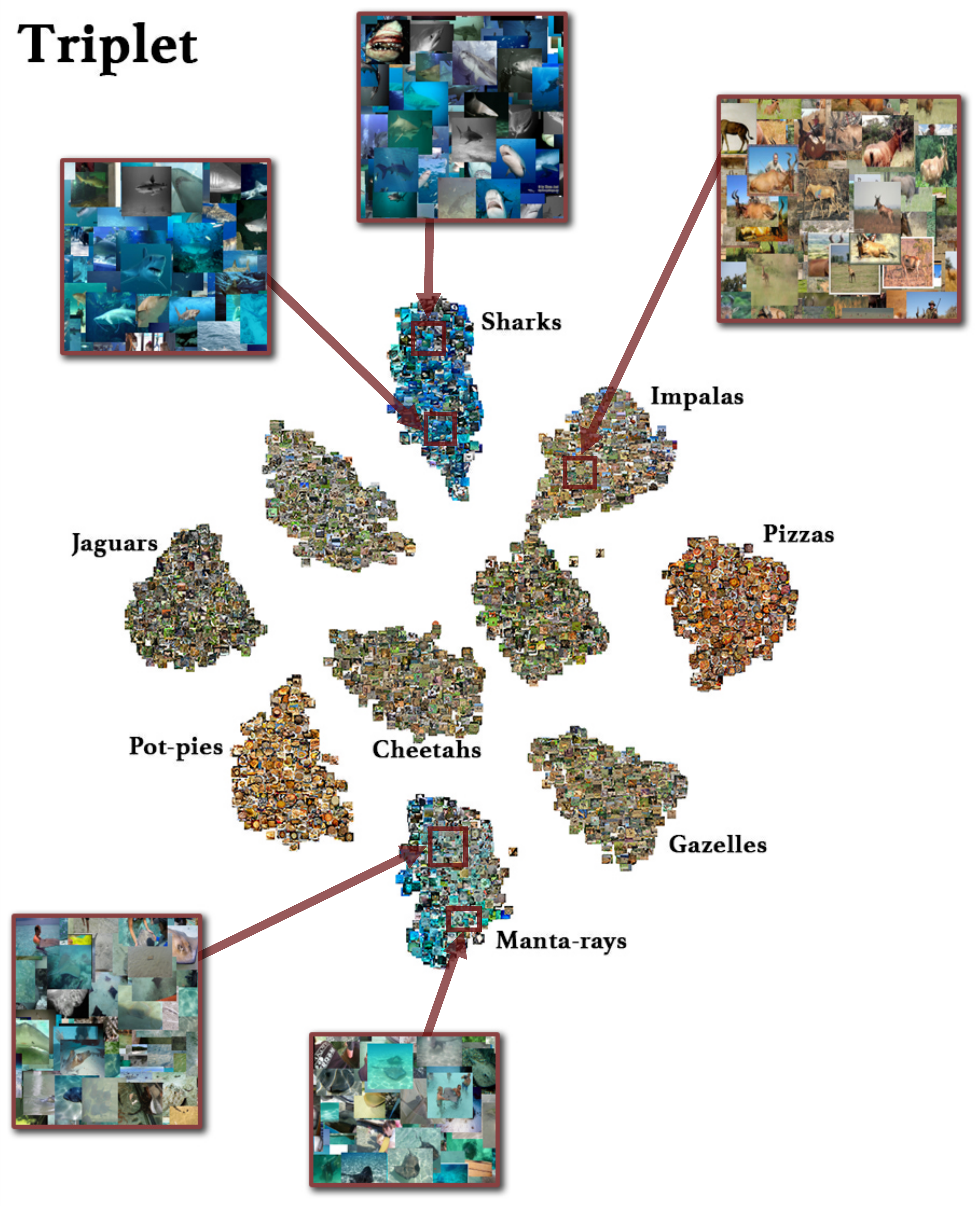}
\end{widepage} 
\end{figure*}
%%%%%%%%%%%%%%%%%%%%%%%%%%%%%%%%%%%%%%%%%%%
\newpage
\section{Hyperparameter Tuning Specifications and Optimal Configurations}\label{app:optimal}
Here we describe in detail the hyperparameter search setups for the different experiments, and the optimal configuration for each.

For all models, we tune optimization hyperparameters consisting of learning rate and its annealing factor which we apply every epoch. We fix the momentum as 0.9 for all experiments. For the smaller datasets, we refresh our index every epoch, and for ImageNet Attributes every 1000 iterations.

For \modelname{}, we additionally tune the separation margin $\alpha$, the number of nearest clusters per minibatch $M$, the number of examples per cluster $D$, and the number of clusters per class $K$ which we take to be the same for all classes (the examples per minibatch $MD$ is upper-bounded by 48 due to memory constraints). Note that we permit the choices $M=D=2$, which, as discussed in \ref{relations}, reverts this back to triplet loss: hence, we expect this choice to be discovered if triplet loss is in fact the optimal choice of distance metric learning loss of this class. For triplet loss, we tune the separation margin $\alpha$, the fraction of nearest impostors retrieved in each minibatch and neighbourhood size retrieved for kNN evaluation.

We now specify the optimal hyperparameter configurations for the different datasets and model spaces, as found empirically via random search. The learning rate annealing factor is marked as ``N/A'' for smaller datasets, where we do not anneal the learning rate at all.

\begin{table*}[h]
\scriptsize
\begin{widepage}
\centering%
  \begin{tabular}[t]{llrrrrr}
    \toprule
    {\bf Model} & {\bf Hyperparameter} & {\bf Pet}  & {\bf Flowers} & {\bf Dogs} & {\bf ImageNet Attributes} & {\bf Hierarchy recovery}\\
    \cmidrule(r){1-7}%
    {\bf \modelnameshort{}}  & Learning rate         & 0.00184  &                   0.0240 &             0.00292 &  0.00459 & 0.00177\\ 
    & Annealing factor         & N/A  &                   N/A &             N/A &  0.974 & 0.988\\
               & Gap                   &                  7.18  &                     2.43 &               0.710 & 0.700 & 0.783 \\
               & Global scaling        &                  3.52  &                     14.2 &                3.03 & 6.42 & 2.33 \\
               & Clusters/class        &                     8  &                        1 &                   1 & 2 & 16 \\
    \cmidrule(r){1-7}%
    {\bf Triplet}    & Learning rate         &         0.000598 &                  0.00155 &             0.00293 & 0.00807 & 0.00187 \\
    & Annealing factor         & N/A  &                   N/A &             N/A &   0.966 & 0.995\\
               & Gap                   &                  0.304 &                    0.554 &               0.370 &   0.495 & 0.556 \\
               & Nearest impostor fraction  &      0.184 &                    0.129 &             0.00713 &  0.0700 & 0.0424 \\
               & Neighbourhood size   &                     128 &                      128 &                 128 &  128 & 128\\
   \bottomrule%
   \newline

\end{tabular}
\caption{Optimal hyperparameter configurations for the different datasets and model spaces.}
\label{fig:optimal_hp}
\end{widepage}
\end{table*}
%%%%%%%%%%%%%%%%%%%%%%%%%%%%%%%%%%%%%%%%%%%

\section{Specifications for ImageNet Attributes Dataset}\label{app:im_attr}
To curate this dataset, we first matched the annotated examples in the Object Attributes dataset \citep{RussakovskyECCV10} to examples in the training set of ImageNet. The ImageNet Attributes training and validation sets then comprise all examples of all classes for which annotated examples exist.

Below we list these classes. 
\begin{lstlisting}[tabsize=1,basicstyle=\scriptsize\ttfamily]
n01693334, n01773549, n01773797, n01796340, n01872401, n01873310, n01882714, n01883070, 
n02071294, n02074367, n02088238, n02088364, n02088466, n02088632, n02090379, n02091134, 
n02091635, n02092002, n02096294, n02100583, n02100735, n02101556, n02102480, n02104029, 
n02104365, n02105056, n02105162, n02105251, n02105505, n02106030, n02109047, n02109525, 
n02110806, n02110958, n02112350, n02115913, n02119789, n02123045, n02123394, n02124075, 
n02125311, n02128925, n02129165, n02130308, n02326432, n02342885, n02361337, n02391049, 
n02410509, n02422106, n02422699, n02423022, n02441942, n02442845, n02443114, n02443484, 
n02444819, n02445715, n02447366, n02480495, n02480855, n02481823, n02483708, n02484975, 
n02486261, n02486410, n02487347, n02488291, n02488702, n02500267, n02509815, n02536864,
n02802426, n02808440, n02910353, n03249569, n03325584, n03721384, n03977966, n03982430, 
n04118776, n04228054, n04447861, n07615774, n07745940, n07873807, n07875152, n07880968, 
n11939491, n12267677
\end{lstlisting}
% %%%%%%%%%%%%%%%%%%%%%%%%%%%%%%%%%%%%%%%%%
\end{appendices}

\bibliographystyle{iclr2016_conference}
\bibliography{dml_arxiv}

\begin{thebibliography}{31}
\providecommand{\natexlab}[1]{#1}
\providecommand{\url}[1]{\texttt{#1}}
\expandafter\ifx\csname urlstyle\endcsname\relax
  \providecommand{\doi}[1]{doi: #1}\else
  \providecommand{\doi}{doi: \begingroup \urlstyle{rm}\Url}\fi

\bibitem[Angelova \& Long(2014)Angelova and Long]{DBLP:conf/wacv/AngelovaL14}
Angelova, Anelia and Long, Philip~M.
\newblock Benchmarking large-scale fine-grained categorization.
\newblock In \emph{{IEEE} Winter Conference on Applications of Computer Vision,
  Steamboat Springs, CO, USA, March 24-26, 2014}, pp.\  532--539, 2014.
\newblock \doi{10.1109/WACV.2014.6836056}.

\bibitem[Angelova \& Zhu(2013)Angelova and Zhu]{DBLP:conf/cvpr/AngelovaZ13}
Angelova, Anelia and Zhu, Shenghuo.
\newblock Efficient object detection and segmentation for fine-grained
  recognition.
\newblock In \emph{2013 {IEEE} Conference on Computer Vision and Pattern
  Recognition, Portland, OR, USA, June 23-28, 2013}, pp.\  811--818, 2013.
\newblock \doi{10.1109/CVPR.2013.110}.

\bibitem[Arthur \& Vassilvitskii(2007)Arthur and
  Vassilvitskii]{Arthur:2007:KAC:1283383.1283494}
Arthur, David and Vassilvitskii, Sergei.
\newblock K-means++: The advantages of careful seeding.
\newblock In \emph{Proceedings of the Eighteenth Annual ACM-SIAM Symposium on
  Discrete Algorithms}, SODA '07, pp.\  1027--1035, Philadelphia, PA, USA,
  2007. Society for Industrial and Applied Mathematics.
\newblock ISBN 978-0-898716-24-5.

\bibitem[Chopra et~al.(2005)Chopra, Hadsell, and
  LeCun]{Chopra:2005:LSM:1068507.1068961}
Chopra, Sumit, Hadsell, Raia, and LeCun, Yann.
\newblock Learning a similarity metric discriminatively, with application to
  face verification.
\newblock In \emph{Proceedings of the 2005 IEEE Computer Society Conference on
  Computer Vision and Pattern Recognition (CVPR'05) - Volume 1 - Volume 01},
  CVPR '05, pp.\  539--546, Washington, DC, USA, 2005. IEEE Computer Society.
\newblock ISBN 0-7695-2372-2.
\newblock \doi{10.1109/CVPR.2005.202}.

\bibitem[Donahue et~al.(2013)Donahue, Jia, Vinyals, Hoffman, Zhang, Tzeng, and
  Darrell]{Donahue_decaf:a}
Donahue, Jeff, Jia, Yangqing, Vinyals, Oriol, Hoffman, Judy, Zhang, Ning,
  Tzeng, Eric, and Darrell, Trevor.
\newblock Decaf: A deep convolutional activation feature for generic visual
  recognition.
\newblock 2013.

\bibitem[Gavves et~al.(2013)Gavves, Fernando, Snoek, Smeulders, and
  Tuytelaars]{GavvesICCV2013}
Gavves, E., Fernando, B., Snoek, C. G.~M., Smeulders, A. W.~M., and Tuytelaars,
  T.
\newblock Fine-grained categorization by alignments.
\newblock In \emph{IEEE International Conference on Computer Vision}, 2013.

\bibitem[Gavves et~al.(2015)Gavves, Fernando, Snoek, Smeulders, and
  Tuytelaars]{gavves2015}
Gavves, Efstratios, Fernando, Basura, Snoek, CeesG.M., Smeulders, ArnoldW.M.,
  and Tuytelaars, Tinne.
\newblock Local alignments for fine-grained categorization.
\newblock \emph{International Journal of Computer Vision}, 111\penalty0
  (2):\penalty0 191--212, 2015.
\newblock ISSN 0920-5691.
\newblock \doi{10.1007/s11263-014-0741-5}.

\bibitem[Globerson \& Roweis(2006)Globerson and Roweis]{NIPS2005_2947}
Globerson, Amir and Roweis, Sam~T.
\newblock Metric learning by collapsing classes.
\newblock In Weiss, Y., Sch\"{o}lkopf, B., and Platt, J.C. (eds.),
  \emph{Advances in Neural Information Processing Systems 18}, pp.\  451--458.
  MIT Press, 2006.
\newblock URL
  \url{http://papers.nips.cc/paper/2947-metric-learning-by-collapsing-classes.pdf}.

\bibitem[Goldberger et~al.(2004)Goldberger, Roweis, Hinton, and
  Salakhutdinov]{Goldberger04neighbourhoodcomponents}
Goldberger, Jacob, Roweis, Sam, Hinton, Geoff, and Salakhutdinov, Ruslan.
\newblock Neighbourhood components analysis.
\newblock In \emph{Advances in Neural Information Processing Systems 17}, pp.\
  513--520. MIT Press, 2004.

\bibitem[Hadsell et~al.(2006)Hadsell, Chopra, and
  Lecun]{Hadsell06dimensionalityreduction}
Hadsell, Raia, Chopra, Sumit, and Lecun, Yann.
\newblock Dimensionality reduction by learning an invariant mapping.
\newblock In \emph{In Proc. Computer Vision and Pattern Recognition Conference
  (CVPR’06}. IEEE Press, 2006.

\bibitem[Ioffe \& Szegedy(2015)Ioffe and Szegedy]{DBLP:conf/icml/IoffeS15}
Ioffe, Sergey and Szegedy, Christian.
\newblock Batch normalization: Accelerating deep network training by reducing
  internal covariate shift.
\newblock In \emph{Proceedings of the 32nd International Conference on Machine
  Learning, {ICML} 2015, Lille, France, 6-11 July 2015}, pp.\  448--456, 2015.

\bibitem[Khosla et~al.(2011)Khosla, Jayadevaprakash, Yao, and
  Fei-Fei]{KhoslaYaoJayadevaprakashFeiFei_FGVC2011}
Khosla, Aditya, Jayadevaprakash, Nityananda, Yao, Bangpeng, and Fei-Fei, Li.
\newblock Novel dataset for fine-grained image categorization.
\newblock In \emph{First Workshop on Fine-Grained Visual Categorization, IEEE
  Conference on Computer Vision and Pattern Recognition}, Colorado Springs, CO,
  June 2011.

\bibitem[Mensink et~al.(2013)Mensink, Verbeek, Perronnin, and
  Csurka]{mensink13pami}
Mensink, Thomas, Verbeek, Jakob, Perronnin, Florent, and Csurka, Gabriela.
\newblock Distance-based image classification: Generalizing to new classes at
  near zero cost.
\newblock \emph{Transactions on Pattern Analysis and Machine Intelligence
  (PAMI)}, 2013.

\bibitem[Min et~al.(2010)Min, van~der Maaten, Yuan, Bonner, and
  Zhang]{DBLP:conf/icml/MinMYBZ10}
Min, Martin~Renqiang, van~der Maaten, Laurens, Yuan, Zineng, Bonner,
  Anthony~J., and Zhang, Zhaolei.
\newblock Deep supervised t-distributed embedding.
\newblock In \emph{Proceedings of the 27th International Conference on Machine
  Learning (ICML-10), June 21-24, 2010, Haifa, Israel}, pp.\  791--798, 2010.

\bibitem[Murray \& Perronnin(2014)Murray and
  Perronnin]{DBLP:conf/cvpr/MurrayP14}
Murray, Naila and Perronnin, Florent.
\newblock Generalized max pooling.
\newblock In \emph{2014 {IEEE} Conference on Computer Vision and Pattern
  Recognition, {CVPR} 2014, Columbus, OH, USA, June 23-28, 2014}, pp.\
  2473--2480, 2014.
\newblock \doi{10.1109/CVPR.2014.317}.

\bibitem[Nilsback \& Zisserman(2008)Nilsback and Zisserman]{Nilsback08}
Nilsback, M-E. and Zisserman, A.
\newblock Automated flower classification over a large number of classes.
\newblock In \emph{Proceedings of the Indian Conference on Computer Vision,
  Graphics and Image Processing}, Dec 2008.

\bibitem[Norouzi et~al.(2012)Norouzi, Fleet, and Salakhutdinov]{NIPS2012_4808}
Norouzi, Mohammad, Fleet, David, and Salakhutdinov, Ruslan~R.
\newblock Hamming distance metric learning.
\newblock In Pereira, F., Burges, C.J.C., Bottou, L., and Weinberger, K.Q.
  (eds.), \emph{Advances in Neural Information Processing Systems 25}, pp.\
  1061--1069. Curran Associates, Inc., 2012.

\bibitem[Parkhi et~al.(2012)Parkhi, Vedaldi, Zisserman, and Jawahar]{parkhi12a}
Parkhi, O.~M., Vedaldi, A., Zisserman, A., and Jawahar, C.~V.
\newblock Cats and dogs.
\newblock In \emph{IEEE Conference on Computer Vision and Pattern Recognition},
  2012.

\bibitem[Qian et~al.(2015)Qian, Jin, Zhu, and Lin]{Qian_2015_CVPR}
Qian, Qi, Jin, Rong, Zhu, Shenghuo, and Lin, Yuanqing.
\newblock Fine-grained visual categorization via multi-stage metric learning.
\newblock In \emph{The IEEE Conference on Computer Vision and Pattern
  Recognition (CVPR)}, June 2015.

\bibitem[Russakovsky \& Fei-Fei(2010)Russakovsky and
  Fei-Fei]{RussakovskyECCV10}
Russakovsky, Olga and Fei-Fei, Li.
\newblock Attribute learning in large-scale datasets.
\newblock In \emph{European Conference of Computer Vision (ECCV), International
  Workshop on Parts and Attributes}, 2010.

\bibitem[Russakovsky et~al.(2015)Russakovsky, Deng, Su, Krause, Satheesh, Ma,
  Huang, Karpathy, Khosla, Bernstein, Berg, and Fei-Fei]{ILSVRC15}
Russakovsky, Olga, Deng, Jia, Su, Hao, Krause, Jonathan, Satheesh, Sanjeev, Ma,
  Sean, Huang, Zhiheng, Karpathy, Andrej, Khosla, Aditya, Bernstein, Michael,
  Berg, Alexander~C., and Fei-Fei, Li.
\newblock {ImageNet Large Scale Visual Recognition Challenge}.
\newblock \emph{International Journal of Computer Vision (IJCV)}, pp.\  1--42,
  April 2015.
\newblock \doi{10.1007/s11263-015-0816-y}.

\bibitem[Salakhutdinov \& Hinton(2007)Salakhutdinov and
  Hinton]{AISTATS07_SalakhutdinovH}
Salakhutdinov, Ruslan and Hinton, Geoffrey~E.
\newblock Learning a nonlinear embedding by preserving class neighbourhood
  structure.
\newblock In Meila, Marina and Shen, Xiaotong (eds.), \emph{Proceedings of the
  Eleventh International Conference on Artificial Intelligence and Statistics
  (AISTATS-07)}, volume~2, pp.\  412--419. Journal of Machine Learning Research
  - Proceedings Track, 2007.

\bibitem[Schroff et~al.(2015)Schroff, Kalenichenko, and
  Philbin]{Schroff_2015_CVPR}
Schroff, Florian, Kalenichenko, Dmitry, and Philbin, James.
\newblock Facenet: A unified embedding for face recognition and clustering.
\newblock In \emph{The IEEE Conference on Computer Vision and Pattern
  Recognition (CVPR)}, June 2015.

\bibitem[Sharif~Razavian et~al.(2014)Sharif~Razavian, Azizpour, Sullivan, and
  Carlsson]{Razavian_2014_CVPR_Workshops}
Sharif~Razavian, Ali, Azizpour, Hossein, Sullivan, Josephine, and Carlsson,
  Stefan.
\newblock Cnn features off-the-shelf: An astounding baseline for recognition.
\newblock In \emph{The IEEE Conference on Computer Vision and Pattern
  Recognition (CVPR) Workshops}, June 2014.

\bibitem[Snoek et~al.(2015)Snoek, Rippel, Swersky, Kiros, Satish, Sundaram,
  Patwary, Prabhat, and Adams]{snoek-etal-2015a}
Snoek, Jasper, Rippel, Oren, Swersky, Kevin, Kiros, Ryan, Satish, Nadathur,
  Sundaram, Narayanan, Patwary, Md. Mostofa~Ali, Prabhat, and Adams, Ryan~P.
\newblock Scalable bayesian optimization using deep neural networks.
\newblock In \emph{International Conference on Machine Learning}, 2015.

\bibitem[Szegedy et~al.(2015)Szegedy, Liu, Jia, Sermanet, Reed, Anguelov,
  Erhan, Vanhoucke, and Rabinovich]{43022}
Szegedy, Christian, Liu, Wei, Jia, Yangqing, Sermanet, Pierre, Reed, Scott,
  Anguelov, Dragomir, Erhan, Dumitru, Vanhoucke, Vincent, and Rabinovich,
  Andrew.
\newblock Going deeper with convolutions.
\newblock In \emph{CVPR 2015}, 2015.

\bibitem[van~der Maaten \& Hinton(2008)van~der Maaten and
  Hinton]{maaten2008visualizing}
van~der Maaten, L.J.P. and Hinton, G.E.
\newblock Visualizing high-dimensional data using t-sne.
\newblock 2008.

\bibitem[Verma et~al.(2012)Verma, Mahajan, Sellamanickam, and
  Nair]{verma2012learning}
Verma, Nakul, Mahajan, Dhruv, Sellamanickam, Sundararajan, and Nair, Vinod.
\newblock Learning hierarchical similarity metrics.
\newblock In \emph{Computer Vision and Pattern Recognition (CVPR), 2012 IEEE
  Conference on}, pp.\  2280--2287. IEEE, 2012.

\bibitem[Wang et~al.(2014)Wang, Song, Leung, Rosenberg, Wang, Philbin, Chen,
  and Wu]{wang2014learning}
Wang, Jiang, Song, Yang, Leung, Thomas, Rosenberg, Chuck, Wang, Jingbin,
  Philbin, James, Chen, Bo, and Wu, Ying.
\newblock Learning fine-grained image similarity with deep ranking.
\newblock In \emph{Computer Vision and Pattern Recognition (CVPR), 2014 IEEE
  Conference on}, pp.\  1386--1393. IEEE, 2014.

\bibitem[Weinberger \& Saul(2009)Weinberger and
  Saul]{Weinberger:2009:DML:1577069.1577078}
Weinberger, Kilian~Q. and Saul, Lawrence~K.
\newblock Distance metric learning for large margin nearest neighbor
  classification.
\newblock \emph{J. Mach. Learn. Res.}, 10:\penalty0 207--244, June 2009.
\newblock ISSN 1532-4435.

\bibitem[Xie et~al.(2015)Xie, Yang, Wang, and Lin]{DBLP:conf/cvpr/XieYWL15}
Xie, Saining, Yang, Tianbao, Wang, Xiaoyu, and Lin, Yuanqing.
\newblock Hyper-class augmented and regularized deep learning for fine-grained
  image classification.
\newblock In \emph{{IEEE} Conference on Computer Vision and Pattern
  Recognition, {CVPR} 2015, Boston, MA, USA, June 7-12, 2015}, pp.\
  2645--2654, 2015.
\newblock \doi{10.1109/CVPR.2015.7298880}.

\end{thebibliography}

\end{document}